\documentclass{article}
\usepackage{spconf,amsmath,graphicx}

\usepackage{dblfloatfix}    % To enable figures at the bottom of page
\usepackage{kantlipsum}     % for random text
\usepackage[detect-all]{siunitx}

\graphicspath{{figures/}{pictures/}{images/}{./}} % where to search for the images
\usepackage{hyperref}
\usepackage{graphicx}
\usepackage{subfig}
\usepackage{tabularx}
\usepackage{float}

\title{Separating Overlapping Tissue Layers from Microscopy Images}
\name{Zahra Montazeri\thanks{e-mail: zmontaze@uci.edu} \quad  \quad  M. Gopi \thanks{e-mail:Gopi@ics.uci.edu} }
\address{Department of Computer Science, University of California, Irvine}

\begin{document}
%\ninept
%
\maketitle
\begin{abstract}
Manual preparation of tissue slices for microscopy imaging can introduce tissue tears and overlaps. Typically, further digital processing algorithms such as registration and 3D reconstruction from tissue image stacks cannot handle images with tissue tear/overlap artifacts, and so such images are usually discarded. In this paper, we propose an imaging model and an algorithm to digitally separate overlapping tissue data of mouse brain images into two layers. We show the correctness of our model and the algorithm by comparing our results with the ground truth.    
%Damaged tissues not only contain no information but also they may have wrong data and distract the data scientists. This is a very crucial issue for biologists. This paper proposes an end-to-end pipeline to separate out the overlaid tissues and retrieve the missing data in the overlaid region. Instead of simply discarding these damaged tissues which are the only option at this point, we can extract the missing values using the known adjacent regions. Considering an imaging model for the image capturing process, it is possible to reverse engineer the mechanism and find the closest value that fits the both of the tissues that are overlapping with each other. 
\end{abstract}


\section{Introduction}
\label{sec:intro}

Tissues of mouse brain images sliced for the purpose of neuron tracing and high resolution digital 3D brain reconstruction, are too thin for robust manual handling \cite{peng2008bioimage, kuan2015neuroinformatics}. At the time the tissues are laid on the glass for imaging, there can be tissue tears and tissue overlaps. Images of such tissues are typically discarded before further digital processing of tissue image stacks since almost none of the algorithms down the pipeline will be able to handle such artifacts. In this paper, we propose how missing regions of two overlaid tissue layers can be reconstructed using texture synthesis.\\ 
%
%Bioimage informatics is a new born area requiring biological knowledge such as developing novel image processing in data-intensive problems \cite{peng2008bioimage}. The diagram of brain's wiring is fundamental and important for understanding brain circuit \cite{oh2014mesoscale}. In order to reconstruct these virtual models, frozen mouse brain is cut into serial slices needed for quantitative analysis \cite{kuan2015neuroinformatics}. Tissues are thin and vulnerable to damage in the capturing process. These damages causes missing information and may even yield wrong diagnosis. In Biology, every single piece of information is important and precious. At this point, damaged tissues with such artifacts are typically discarded and recaptured if needed. In this paper, we show how missing regions of a damaged tissue can be retrieved using Texture Synthesis and neighbor information \cite{Efros:1999:TSN:850924.851569}.\\ 

There have been previous works on repairing tissue damage artifacts. However, these earlier works paid attention to some of the other microscopy image artifacts. For instance, \cite{kindle2011semiautomated} explored dust air bubbles and particles. Tissue deformations such as stretching and compaction have been addressed in \cite{ju20063d}. Moreover, blind separation approach has become popular for separation of two independent overlapping images that have completely different components \cite{bell1995information, 4431868}. Obviously, it fails on separating overlapping tissue layers of a single microscopy image that has same components. Also, same approach is used for denoising medical images in \cite{Mustafi2013}. Furthermore, in our recent work \cite{Nitin1,Nitin2}, we developed robust registration algorithm to handle images with tissue tears or with missing tissue data. However, tissue overlap artifact handling is still an open problem and is addressed in this article.

\begin{figure}[tb]
 \centering % avoid the use of \begin{center}...\end{center} and use \centering instead (more compact)
 \includegraphics[width=\columnwidth]{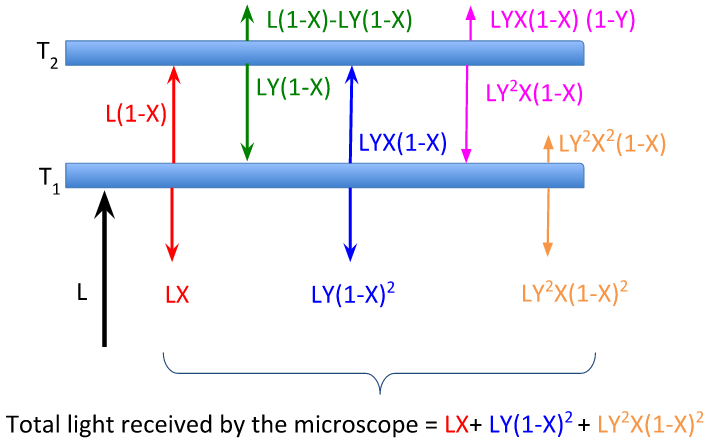}
 \caption{Imaging model with respect to the microscope capturing process. L is the incoming light, $X$ and $Y$ are the reflection coefficients for two overlapping tissue layers $T_1$ and $T_2$, respectively.}
 \label{fig:model}
\end{figure}

\section{Proposed Method}
In this section, we propose a method to separate two overlapping tissues in digital images into two layers. The input to our method, is the section of the digital microscopy image that has two-tissue overlap as well as some part outside the overlap region. Using known information in the non-overlapping neighborhoods of the overlapping region and the imaging model, two individual tissue layers are computed and separated. The outputs of our method are two digitally separated tissue layers.

\subsection{Texture Synthesis} \label{texture synthesis}
the information from adjacent non-overlapping neighborhood of the boundary of the overlaps provide the initialization data to start the optimization process of reconstructing the individual tissue layers. We use one of the hole filling methods \cite{Efros:1999:TSN:850924.851569, darabi2012image}, PatchMatch \cite{Barnes:2009:PAR}, to initialize the two tissue layers. Following the initialization, the content of each tissue layer is iteratively optimized so that a virtual image of the overlap of the two reconstructed layers matches with the real input image of the overlapping region. We also propose a  microscope imaging model in order to capture a virtual image. 

\subsection{Imaging Model} \label{imaging model}
We propose an imaging model to specifically express the overlapping tissue capturing process. Using this model, we virtually capture the digital overlap of the separated output tissues and compare it with the original captured overlaid region to evaluate the results. The imaging model is illustrated in \autoref{fig:model} where L is the incoming light, $X$ and $Y$ are the reflection coefficients for the two overlapping tissue layers $T_1$ and $T_2$, respectively. 
$X$ and $Y$ are initialized using texture synthesis methods discussed in Section \ref{texture synthesis}. They should be updated such that the obtained output $Z$ after applying the imaging model, would be the closest value at that particular pixel in the original overlaid area, cropped from the input. If $Z$ is the total light received by the microscope, then regarding to \autoref{fig:model}:
\begin{equation} \label{Z}
 Z = X + Y(1-X)^2 + XY^2(1-X)^2 
 \end{equation}
 The two assumptions of our imaging model of overlapping tissue is that the intensity of the source light $L$ is one, and the amount of light absorbed by the tissue is zero. In other words, the incoming light is equal to the sum of the reflected and transmitted light by the tissue. It is worth mentioning that we applied the model just for two paths of bouncing, and we assume that the light that is not accounted for is negligible.

\subsection{Regularization} \label{regularization} 
\begin{figure*}[!t]
 \centering % avoid the use of \begin{center}...\end{center} and use \centering instead (more compact)
\begin{tabular}{ccc}

\subfloat[Input image with overlap artifact]{\includegraphics[width=0.3\textwidth, height=3.3cm]{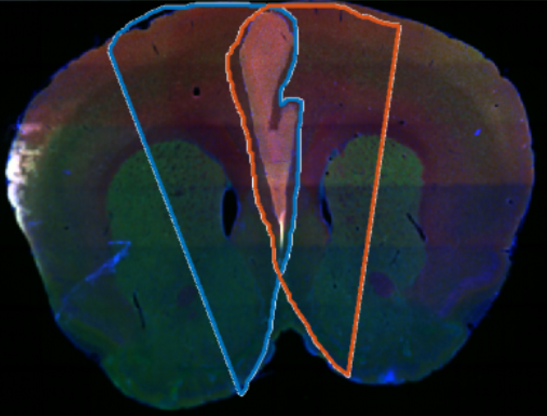}}
&
\subfloat[Original overlaid area $Z'$]{\includegraphics[width=0.3\textwidth, height=3.3cm]{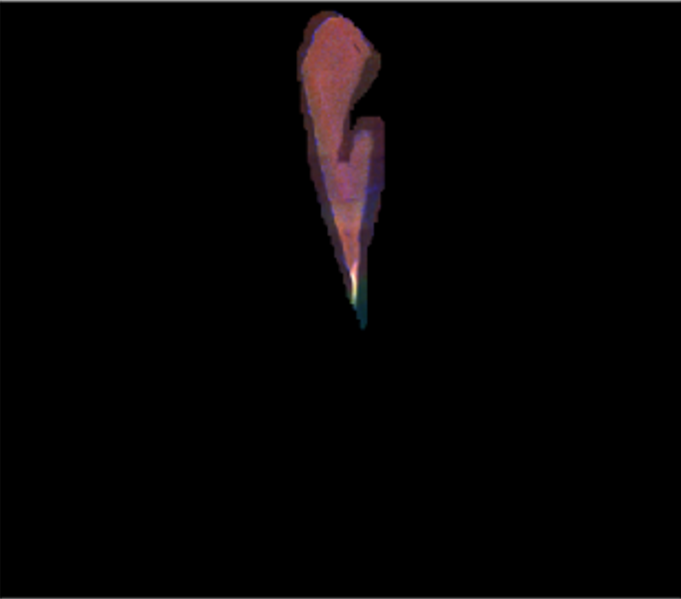}} 
&
\subfloat[Obtained overlaid area $Z$]{\includegraphics[width=0.3\textwidth, height=3.3cm]{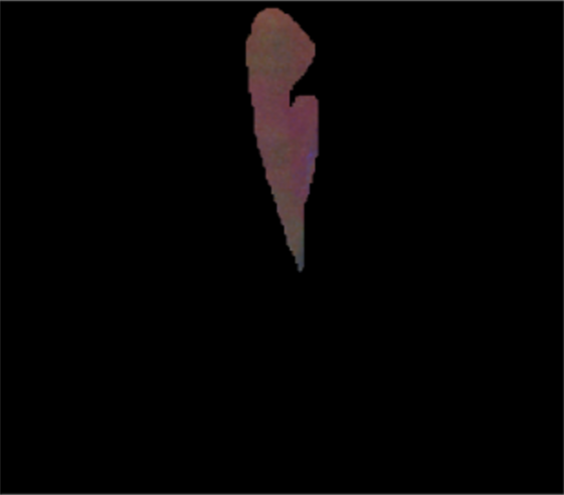}} \\

\end{tabular}
 \caption{The error is defined by the MSE between (b) and (c).}
 \label{fig:results}
\end{figure*}

The initialized $X$ and $Y$ produced in Section \ref{imaging model} can be tuned in order to minimize the error which is defined as the Mean Square Error (MSE) between the obtained virtual imaging of the overlaid region $Z$ from \autoref{Z} and the input original image of the overlaid region $Z'$. $Z$ and $Z'$ are displayed in \autoref{fig:results} (b) and (c). For this purpose, two weights are used to update the current $X$ and $Y$ according to \autoref{equation:regularization}. 

 \begin{equation} \label{equation:regularization}
X = X\times{W_1}  \quad \textrm{and} \quad Y = Y\times{W_2} 
 \end{equation}
 
\begin{figure}[tb]
 \centering % avoid the use of \begin{center}...\end{center} and use \centering instead (more compact)
 \includegraphics[width=\columnwidth]{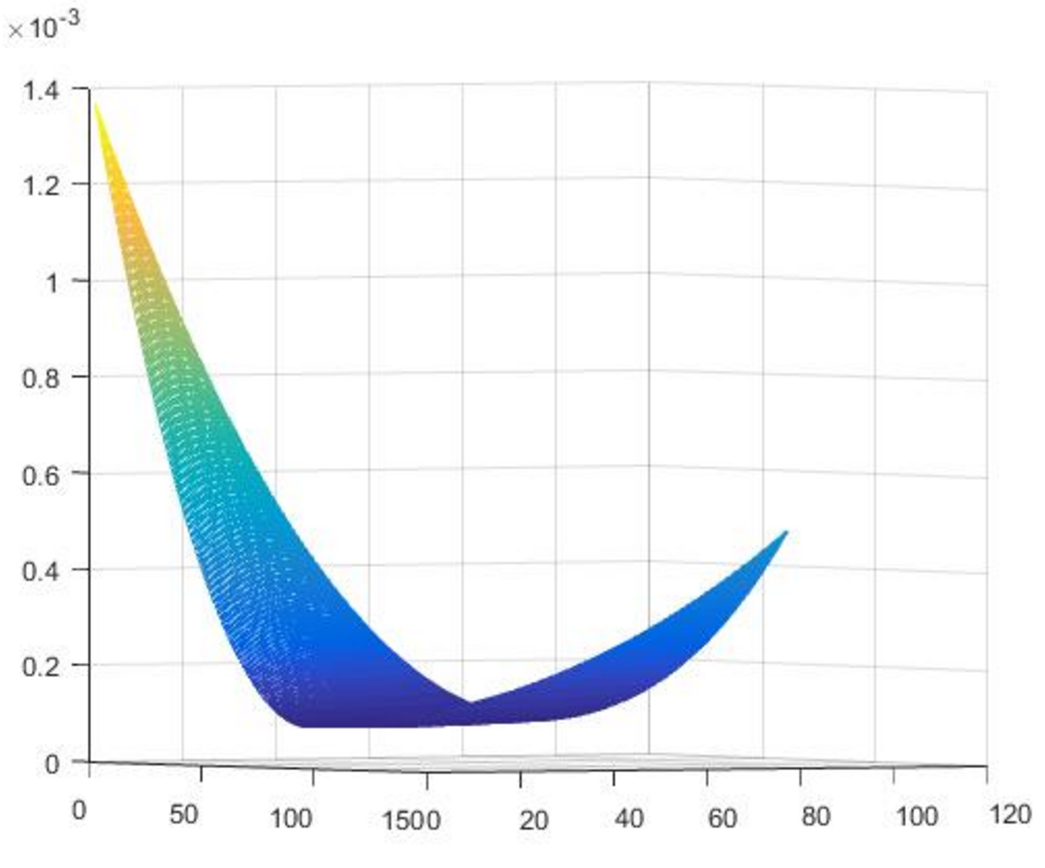}
 \caption{Error surface corresponding to a pair of weights for both engaging tissue layers}
 \label{fig:reg_error}
\end{figure}

The corresponding error $Z-Z'$ is calculated for every possible pair of updated $X$ and $Y$ in terms of all $W_1$ and $W_2$ in the range of \numrange{0}{1} represented in \autoref{fig:reg_error}. The optimum $X$ and $Y$ are obtained using gradient descent which is discussed in Section \ref{gradient descent}.

\subsubsection{Gradient Descent Optimization} \label{gradient descent} 
In order to find the optimum $X$ and $Y$, we use gradient descent with the objective function $G$ visualized in \autoref{fig:reg_error}. The gradient of $G$ with respect to $X$ and $Y$ are calculated as follows according to \autoref{Z}.
%%%%%%%%%%%%
The $X$ and $Y$ values are updated until the error is within the threshold $\epsilon$.
\[G = min \sum_{i=1}^{n}\sum_{j=1}^{m}(Z_{ij} - Z'_{ij})^2\]	
\[\nabla {G} = [\frac{\partial{G}}{\partial{x_{11}}}, \frac{\partial{G}}{\partial{x_{12}}},\ldots ,\frac{\partial{G}}{\partial{x_{nm}}}, 
\frac{\partial{G}}{\partial{y_{11}}}, \frac{\partial{G}}{\partial{y_{12}}}, \ldots, \frac{ \partial{G}}{\partial{y_{nm}}}]\]
\[\partial{G}/\partial{x_{ij}} = 1 - 2 (1-x_{ij})y_{ij} - 2x_{ij}(1-x_{ij})y_{ij}^2+(1-x_{ij})^2y_{ij}^2\]
\[\partial{G}/\partial{y_{ij}} = (1-x_{ij})^2 +2x_{ij}(1-x_{ij})^2y_{ij}\]
\[X_t = X_{t-1} - \alpha\nabla {G} \quad \textrm{and} \quad Y_t = Y_{t-1} - \alpha\nabla {G}\]
\[|G(X_{t}) - G(X_{t-1})| < \epsilon \quad \textrm{and} \quad |G(Y_{t}) - G(Y_{t-1})| < \epsilon\]

\begin{figure*}[!t]
\centering
\begin{tabular}{ccc}

\subfloat{\includegraphics[width=0.3\textwidth, height=3.3cm]{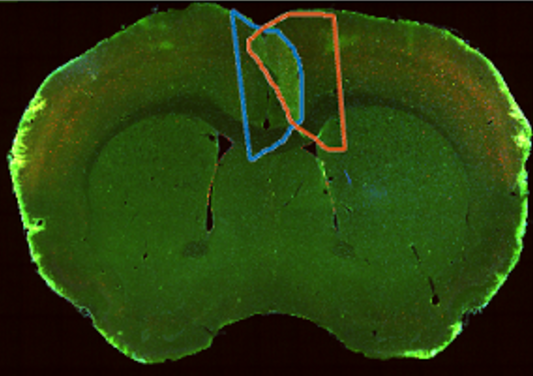}}  
&
\subfloat{\includegraphics[width=0.3\textwidth, height=3.3cm]{user}}
&
\subfloat{\includegraphics[width=0.3\textwidth, height=3.3cm]{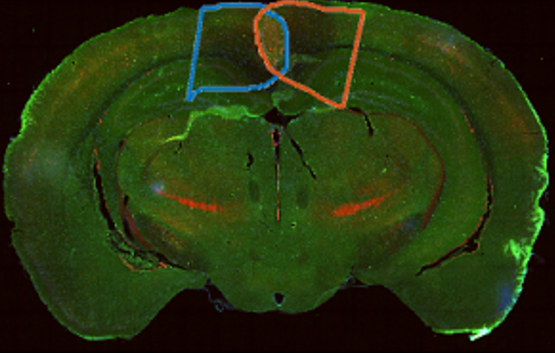}} \\

\subfloat{\includegraphics[width=0.3\textwidth, height=3.3cm]{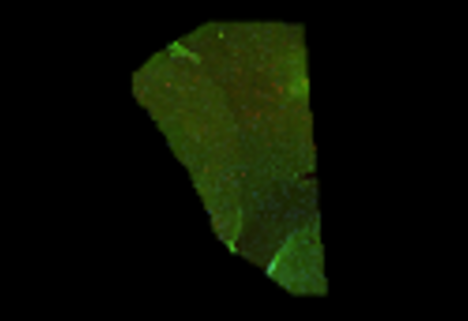}} 
&
\subfloat{\includegraphics[width=0.3\textwidth, height=3.3cm]{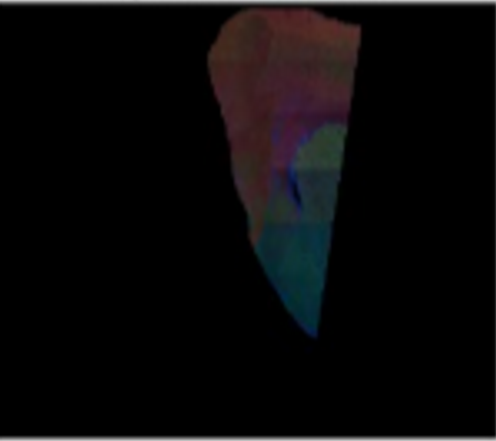}} 
&
\subfloat{\includegraphics[width=0.3\textwidth, height=3.3cm]{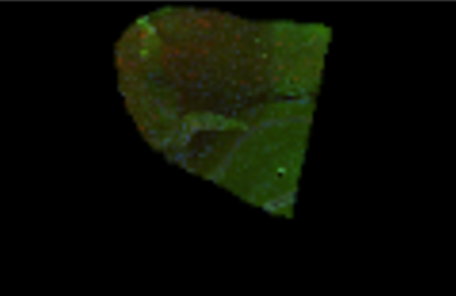}} \\

\subfloat{\includegraphics[width=0.3\textwidth, height=3.3cm]{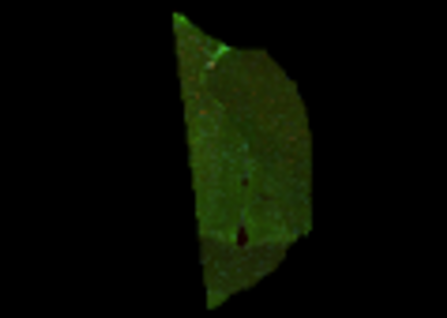}} 
&
\subfloat{\includegraphics[width=0.3\textwidth, height=3.3cm]{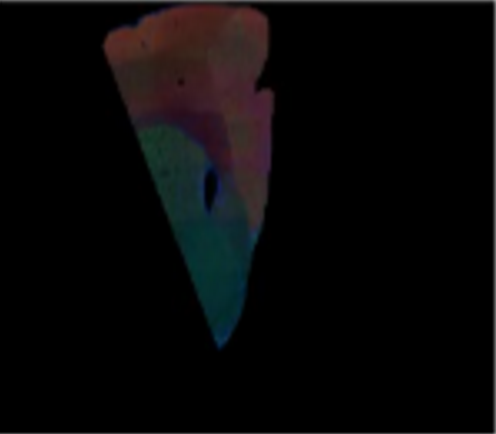}} 
&
\subfloat{\includegraphics[width=0.3\textwidth, height=3.3cm]{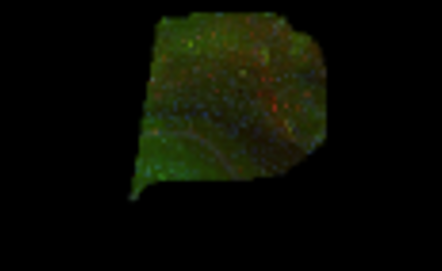}} \\

\subfloat{\includegraphics[width=0.3\textwidth, height=3.3cm]{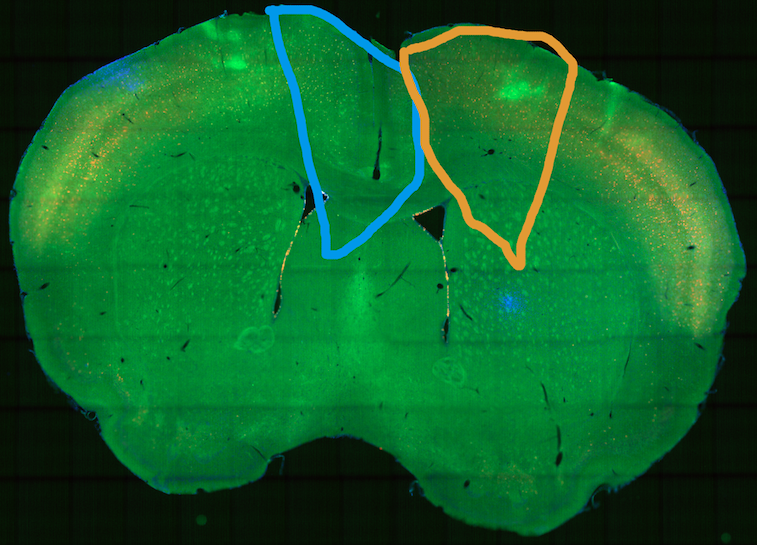}} 
&
\subfloat{\includegraphics[width=0.3\textwidth, height=3.3cm]{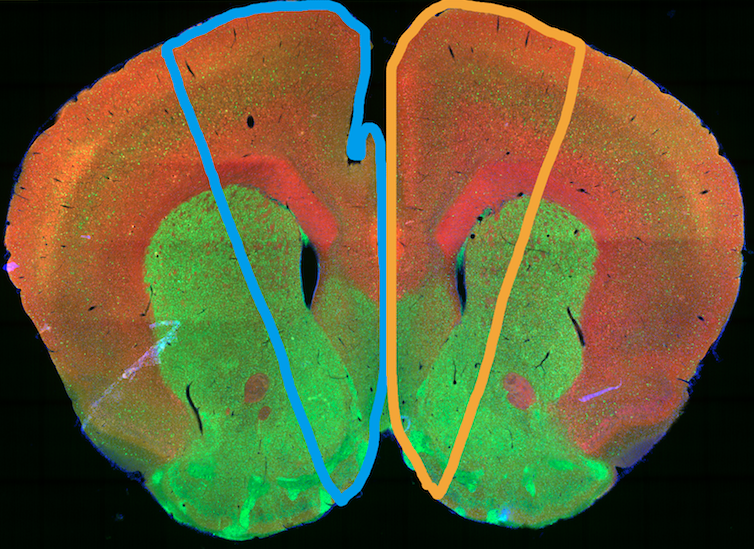}} 
&
\subfloat{\includegraphics[width=0.3\textwidth, height=3.3cm]{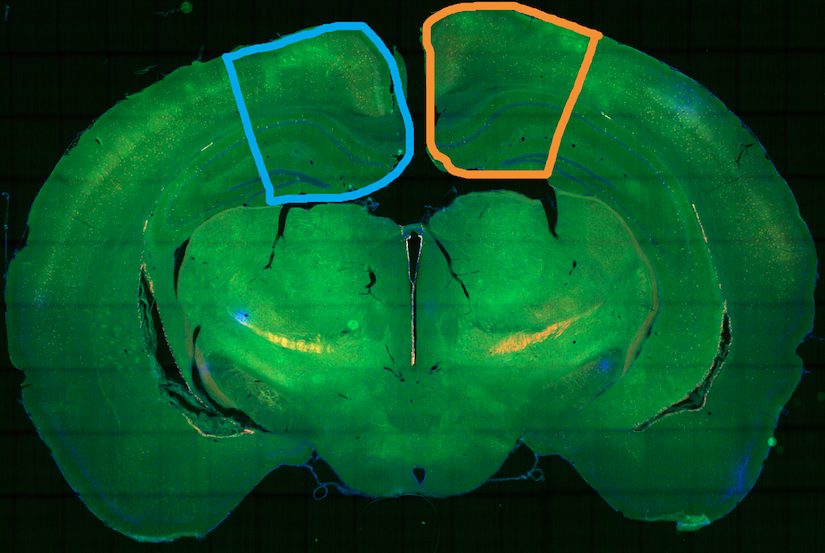}} \\

\end{tabular}
 \caption{A sample of three mouse brain slices with overlaid artifact and the overlapping contours depicted by the user are shown in the first row. Cropped recovered region for the right and left side of the damaged tissue within their contours are shown in the second and third row, respectively. Last row represents the separated corresponding contours on the ground truth images.}
 \label{fig:samples}
\end{figure*}

\begin{figure*}[!b]
 \centering % avoid the use of \begin{center}...\end{center} and use \centering instead (more compact)
\begin{tabular}{cccc}

\subfloat[Obtained right tissue layer]{\includegraphics[width=0.23\textwidth, height=3.2cm]{b1}}  
&
\subfloat[Ground truth for the right tissue]{\includegraphics[width=0.23\textwidth, height=3.2cm]{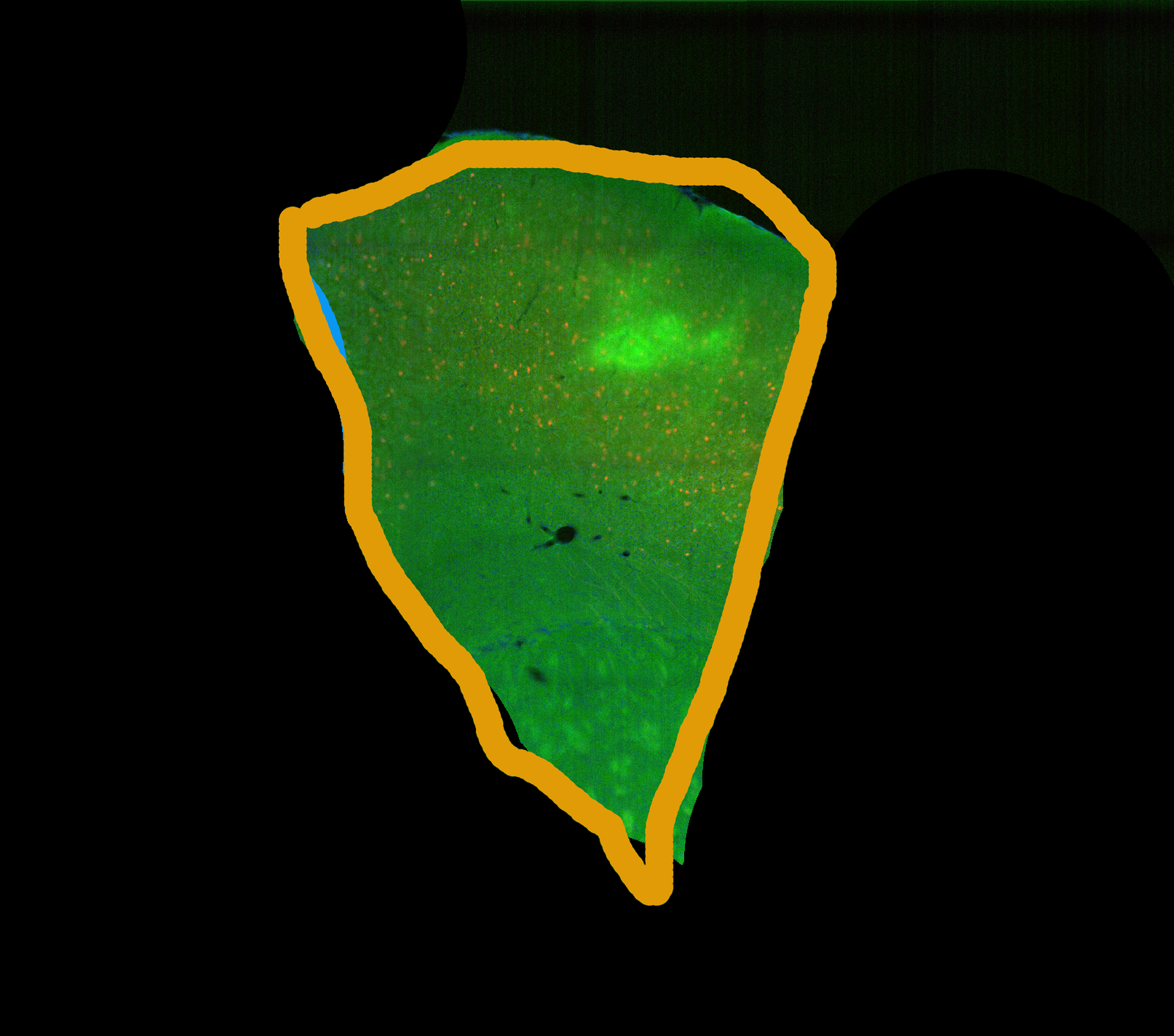}}
&
\subfloat[Obtained left tissue layer]{\includegraphics[width=0.23\textwidth, height=3.2cm]{c1}} 
&
\subfloat[Ground truth for the left tissue]{\includegraphics[width=0.23\textwidth, height=3.2cm]{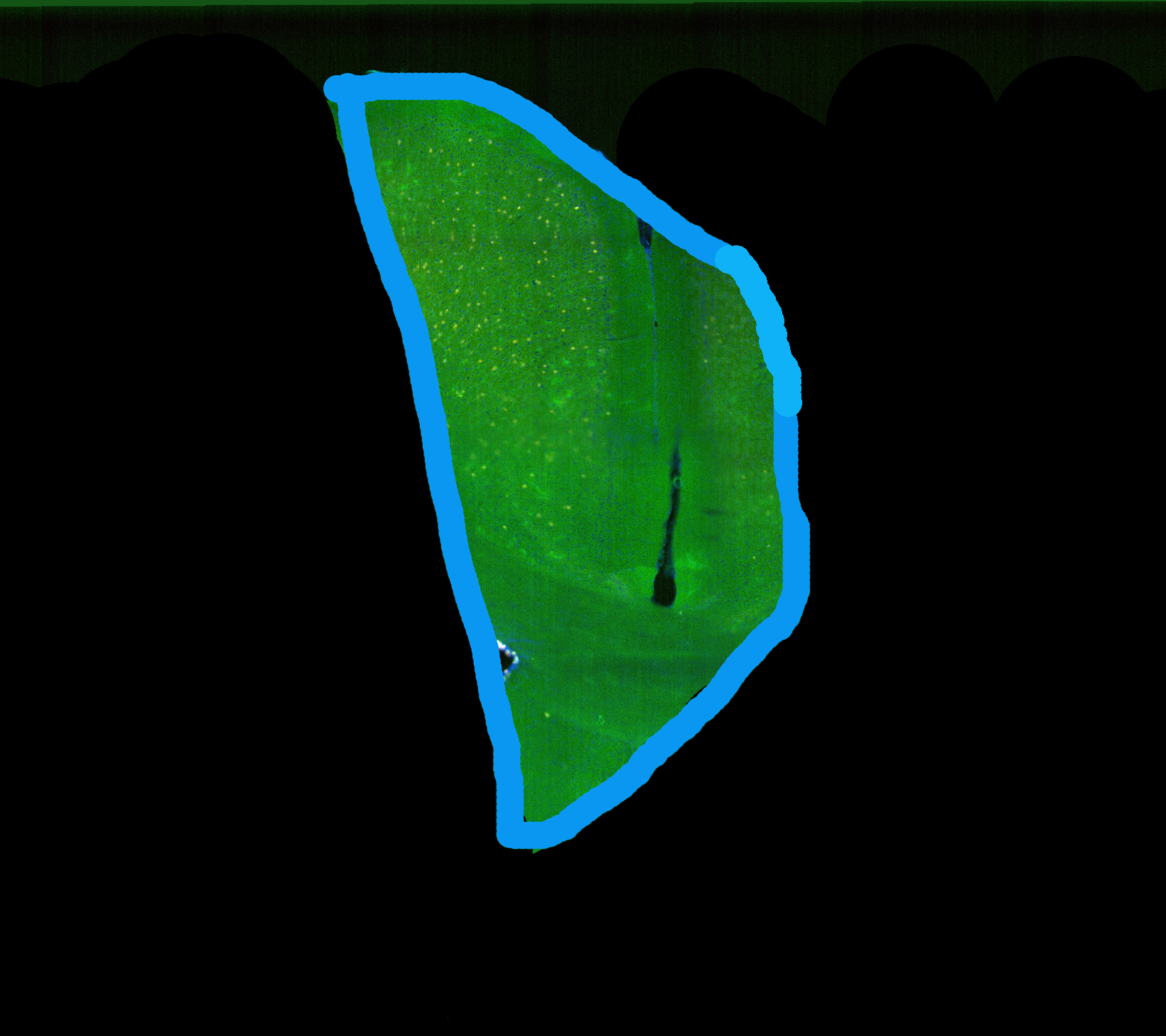}} \\

\end{tabular}
 \caption{Visual evaluation of the reconstructed tissue layers with the corresponding contour in the ground truth image achieved by recapturing damaged slices for the first sample in Figure 4.}
 \label{fig:eval}
\end{figure*}

\subsection{Implementation and Experiment}
The code is implemented in Matlab and covers all the aforementioned steps in this section. 
Here is an overview of analysis pipeline illustrated in Figures \ref{fig:results} and \ref{fig:samples}. The input to our method presented in \autoref{fig:results} image (a). The overlaid region is shown brighter, which is expected and predicted by the proposed imaging model. In the middle image the original overlaid region of the two tissue layers is depicted. This is used as the ground truth for regularization section as well as optimization step to be compared with the obtained $Z$ displayed in image (c). As it is discussed in Section \ref{texture synthesis}, first, Patch-Match algorithm applied. Then, the regularization step is performed addressed in Section \ref{regularization}. In \autoref{fig:reg_error}, the error for different pair of weights are visualized in 3D. The chart illustrates how the error is changed over the different values. The weights yielding the minimum error are picked for the rest of the pipeline using gradient descent optimization, covered in Section \ref{gradient descent}. The iteration stops when the error between the computed and the goal drops below a defined tolerance. The second and third row of \autoref{fig:samples} demonstrate the final results after separating the overlaid tissue layers.

\section{Results and Evaluation} \label{results}

Our method provides two separated tissue layers as an output after employing the aforementioned steps. The final results of three samples with overlapping artifact are shown in \autoref{fig:samples}.  In the first row. Results for the reconstructed right and left side of the damaged tissue within their contours are shown in the second and third row, respectively.
Most importantly, just for evaluating the proposed microscopy image capture model, we physically separated the tissue slices and recaptured the microscopy images of these separated layers, and compared the new image with the layers of tissue separated by our algorithm. This is represented in the last row. Due to new stretching and contraction of tissues, the new slice will not have pixel to pixel correspondence with the old slice that has overlapping tissue layers. So the evaluation of the correctness of our result is only visual.\\
\autoref{fig:eval} expresses the comparison between the obtained results for the two engaging tissue layers and the ground truth which is achieved by recapturing the slice and registering the corresponding regions. The results are from the first sample shown in \autoref{fig:samples}. Recapturing process produces different illumination and a slight changes in orientation due to manual interaction with the slice. Therefore, the ground truth image looks brighter and stretched than the obtained result. According to the results we tested for many samples, it is shown that the overall behavior of the texture has been reconstructed perfectly.

%This application provides two separated tissues as an output after employing the aforementioned steps. Since, the proposed method is novel, we could not directly compare the results with the existing techniques. So, we assume the imaging model is correct and therefore, we evaluate the rest of the pipeline. The results are compared with two hole-filling techniques. New assignments are provided by each technique for the overlaid region using known neighbors, then the imaging model is applied to make the results comparable with the ground truth. The final result is depicted on \autoref{fig:eval}. It represents the overlaid region after applying the mentioned two texture synthesis techniques comparing with the proposed method. The ground truth is the cropped image of the input. The error for Patch-Match and Content-Aware technique are $6.6\times10^{-4}$ and $5.5\times10^{-4}$, respectively, and the observed error for the proposed method is $5.9\times10^{-4}$. Surprisingly, content-aware technique behaves better than our method in terms of error value, quantitatively. However, it does not look reasonable in \autoref{fig:eval} qualitatively. This happens because it deals with the average of the region and that is why the error Mean-Square error is less but visually, it looks worse. In the second column of the figure, the result after employing Patch-Match is represented as we expected. The figure illustrates the proposed method outperforms Patch-Match visually.

\section{Future Work and Conclusion}
One of the important plausible works is to extend the method to support the tissues that are folded multiple times. For this purpose, a new imaging model must be proposed depending on the number of overlapping tissues. Then, we can start with the initialization using texture synthesis as before, but one side of known region must be used for estimating the missing data each time. For the next unknown region, the obtained information for the previous unknown area can be used and similarly, the value for the following region will be needed for obtaining the further layers. Obviously, the more folded layers, the less accurate our method would performs.\\
Another potential direction of future research would be to estimate two indicated contours automatically instead of the user manipulation that we already have. As it mentioned in Section \ref{results}, the overlaid region is the brightest area in the tissue which is perceivable by considering the proposed imaging model. Therefore, the overlaid region can be detected automatically. Adequate known neighbor regions are also needed for both sides of the tissue. Having these three regions, one for the bright overlaid region, another two for depicting two neighborhoods, we are all set to pursue the rest of the pipeline. 
\section{Conclusion}
This article proposes a microscopy imaging model for two layer tissue overlap and an efficient end-to-end pipeline for digital separation of overlaid tissues in microscopy images. We have demonstrated the successful results of the two separated tissues after applying our model comparing with the ground truth. We first start with a hole-filling technique, described in Section \ref{texture synthesis}, as an initialization for the two tissue layers. Then we apply regularization step, discussed in Section \ref{regularization}. Then, an optimization technique is applied to obtain the optimum values for both sides with respect to the defined objective function. This step has been described in Section \ref{gradient descent}. Section \ref{results} evaluates the results by comparing them with the ground truth. 

\section{Acknowledgement} 
The authors would like to thank Xiaoxiao Lin for providing the mouse brain images.

\bibliographystyle{IEEEbib}
\bibliography{template}

\begin{thebibliography}{10}

\bibitem{peng2008bioimage}
Hanchuan Peng,
\newblock ``Bioimage informatics: a new area of engineering biology,''
\newblock {\em Bioinformatics}, vol. 24, no. 17, pp. 1827--1836, 2008.

\bibitem{kuan2015neuroinformatics}
Leonard Kuan, Yang Li, Chris Lau, David Feng, Amy Bernard, Susan~M Sunkin,
  Hongkui Zeng, Chinh Dang, Michael Hawrylycz, and Lydia Ng,
\newblock ``Neuroinformatics of the allen mouse brain connectivity atlas,''
\newblock {\em Methods}, vol. 73, pp. 4--17, 2015.

\bibitem{kindle2011semiautomated}
LM~Kindle, IA~Kakadiaris, T~Ju, and JP~Carson,
\newblock ``A semiautomated approach for artefact removal in serial tissue
  cryosections,''
\newblock {\em Journal of microscopy}, vol. 241, no. 2, pp. 200--206, 2011.

\bibitem{ju20063d}
Tao Ju, Joe Warren, James Carson, Musodiq Bello, Ioannis Kakadiaris, Wah Chiu,
  Christina Thaller, and Gregor Eichele,
\newblock ``3d volume reconstruction of a mouse brain from histological
  sections using warp filtering,''
\newblock {\em Journal of Neuroscience Methods}, vol. 156, no. 1, pp. 84--100,
  2006.

\bibitem{bell1995information}
Anthony~J Bell and Terrence~J Sejnowski,
\newblock ``An information-maximization approach to blind separation and blind
  deconvolution,''
\newblock {\em Neural computation}, vol. 7, no. 6, pp. 1129--1159, 1995.

\bibitem{4431868}
E.~Be'ery and A.~Yeredor,
\newblock ``Blind separation of superimposed shifted images using parameterized
  joint diagonalization,''
\newblock {\em IEEE Transactions on Image Processing}, vol. 17, no. 3, pp.
  340--353, March 2008.

\bibitem{Mustafi2013}
A.~Mustafi and S.K. Ghorai,
\newblock ``A novel blind source separation technique using fractional fourier
  transform for denoising medical images,''
\newblock {\em Optik - International Journal for Light and Electron Optics},
  vol. 124, no. 3, pp. 265--271, 2013.

\bibitem{Nitin1}
Nitin Agarwal, Xiangmin Xu, and M.~Gopi,
\newblock ``Robust registration of mouse brain slices with severe histological
  artifacts,''
\newblock in {\em Proceedings of the Tenth Indian Conference on Computer
  Vision, Graphics and Image Processing}. 2016, ICVGIP '16, p.~10, ACM.

\bibitem{Nitin2}
Nitin Agarwal, Xiangmin Xu, and M.~Gopi,
\newblock ``Automatic detection of histological artifacts in mouse brain slice
  images,''
\newblock in {\em International MICCAI Workshop on Medical Computer Vision}.
  Springer, 2016, p. TBD.

\bibitem{Efros:1999:TSN:850924.851569}
Alexei~A. Efros and Thomas~K. Leung,
\newblock ``Texture synthesis by non-parametric sampling,''
\newblock in {\em Proceedings of the International Conference on Computer
  Vision-Volume 2 - Volume 2}, Washington, DC, USA, 1999, ICCV '99, pp. 1033--,
  IEEE Computer Society.

\bibitem{darabi2012image}
Soheil Darabi, Eli Shechtman, Connelly Barnes, Dan~B Goldman, and Pradeep Sen,
\newblock ``Image melding: Combining inconsistent images using patch-based
  synthesis.,''
\newblock {\em ACM Trans. Graph.}, vol. 31, no. 4, pp. 82--1, 2012.

\bibitem{Barnes:2009:PAR}
Connelly Barnes, Eli Shechtman, Adam Finkelstein, and Dan~B Goldman,
\newblock ``{PatchMatch}: A randomized correspondence algorithm for structural
  image editing,''
\newblock {\em ACM Transactions on Graphics (Proc. SIGGRAPH)}, vol. 28, no. 3,
  Aug. 2009.

\end{thebibliography}

\end{document}